\documentclass[preprint,12pt]{elsarticle}

\usepackage{amssymb}
\usepackage[utf8]{inputenc}
\usepackage[T1]{fontenc}
\usepackage{url,lineno,microtype}
\usepackage[onehalfspacing]{setspace}
\usepackage{booktabs}
\usepackage{hyperref}
\usepackage{multirow}
\usepackage{threeparttable}
\usepackage{subcaption}
\usepackage{color}
\usepackage{times}
\usepackage{epsfig}
\usepackage{graphicx}
\usepackage{amsmath}
\usepackage{amsthm}
\usepackage{amssymb}
\usepackage{algorithm}
\usepackage{algorithmic}
\usepackage{makecell}
\definecolor{reviewerblue}{RGB}{120,200,240}        % blue
\definecolor{reviewerred}{RGB}{250,10,10}         % red
\definecolor{reviewergreen}{RGB}{0,120,0}       % green
\definecolor{reviewerorange}{RGB}{255,128,0}    % orange
\definecolor{reviewerpurple}{RGB}{75,0,128}     % purple
\definecolor{reviewergrey}{RGB}{128,128,128}     % grey

\journal{}

\begin{document}

\begin{frontmatter}



\title{Spiking Neural Networks with Consistent Mapping Relations Allow High-Accuracy Inference}


\author[inst1,inst2]{Yang Li \corref{cor0}}
\ead{liyang2019@ia.ac.cn}
\author[inst1,inst2]{Xiang He \corref{cor0}}
\ead{hexiang2021@ia.ac.cn}
\author[inst1,inst3]{Qingqun Kong}
\ead{qingqun.kong@ia.ac.cn}
\author[inst1,inst2,inst3,inst4]{Yi Zeng\corref{cor1}}
\ead{yi.zeng@ia.ac.cn}
\cortext[cor0]{These authors contributed equally to this work.}
\cortext[cor1]{Corresponding author}

\affiliation[inst1]{organization={Brain-inspired Cognitive Intelligence Lab, Institute of Automation, Chinese Academy of Sciences (CAS)},%Department and Organization
            city={Beijing},
            country={China}}
            
\affiliation[inst2]{organization={School of Artificial Intelligence, University of Chinese Academy of Sciences},%Department and Organization
            city={Beijing},
            country={China}}

\affiliation[inst3]{organization={School of Future Technology, University of Chinese Academy of Sciences},%Department and Organization
            city={Beijing},
            country={China}}

\affiliation[inst4]{organization={Center for Excellence in Brain Science and Intelligence Technology, CAS},%Department and Organization
            city={Shanghai},
            country={China}}

\begin{abstract}
%% Text of abstract \textcolor{\reviewerred}{}
Spike-based neuromorphic hardware has demonstrated substantial potential in low energy consumption and efficient inference. However, the direct training of deep spiking neural networks is challenging, and conversion-based methods still require substantial time delay owing to unresolved conversion errors. We determine that the primary source of the conversion errors stems from the inconsistency between the mapping relationship of traditional activation functions and the input-output dynamics of spike neurons. To counter this, we introduce the Consistent ANN-SNN Conversion (CASC) framework. It includes the Consistent IF (CIF) neuron model, specifically contrived to minimize the influence of the stable point's upper bound, and the wake-sleep conversion (WSC) method, synergistically ensuring the uniformity of neuron behavior. This method theoretically achieves a loss-free conversion, markedly diminishing time delays and improving inference performance in extensive classification and object detection tasks. Our approach offers a viable pathway toward more efficient and effective neuromorphic systems.
\end{abstract}


% \begin{highlights}
% \end{highlights}

\begin{keyword}
Spiking neural network \sep conversion \sep consistency \sep object detection
% %% PACS codes here, in the form: \PACS code \sep code
% %% MSC codes here, in the form: \MSC code \sep code
% %% or \MSC[2008] code \sep code (2000 is the default)
\end{keyword}

\end{frontmatter}

% \linenumbers

\section{Introduction}

Artificial intelligence (AI), represented by deep neural networks, has achieved numerous remarkable successes despite being accompanied by colossal energy consumption, which severely limits AI development, particularly in edge devices. Spiking neural network (SNN), due to its event computing attributes, which when combined with neuromorphic hardware such as TrueNorth demonstrates great potential for efficient inference \cite{roy2019towards}, has gained substantial attention. However, owing to the non-differentiable nature of the spiking processes, training high-performance SNNs becomes a challenge that hampers SNN development.

Spiking neurons, with their intricate dynamical attributes, improve the biomimetic essence and credibility of SNNs. This is particularly evident in the context of spiking neural P systems, as revealed by several studies \cite{zhang2014optimization,zhang2022layered}. Moreover, techniques such as Hebbian learning \cite{krotov2019unsupervised}, spike-timing-dependent plasticity (STDP) \cite{srinivasan2017spike}, and enzymes \cite{zhang2022enzymatic} have been vital in SNN training. Even with these advancements, the prolonged simulation time and suboptimal efficiency of SNNs still fall short of those of traditional ANNs.
Thus, to train SNNs directly, smooth gradients are employed to replace the delta process\cite{wu2018spatio,shrestha2018slayer,wei2023temporal,zhang2021rectified}. Furthermore, the surrogate gradient (SG) method can fully utilize the ability of SNNs to explore spatiotemporal information representations, and therefore, improve SNN performance. Some recent studies have revealed that in large-scale datasets, SNNs can perform as well as ANNs\cite{deng2022temporal}. Nevertheless, training SNNs necessitates storage activation and gradient values in multiple time steps, which increases the computational burden.

\begin{figure}[t]
	\centering
	\includegraphics[scale=0.48]{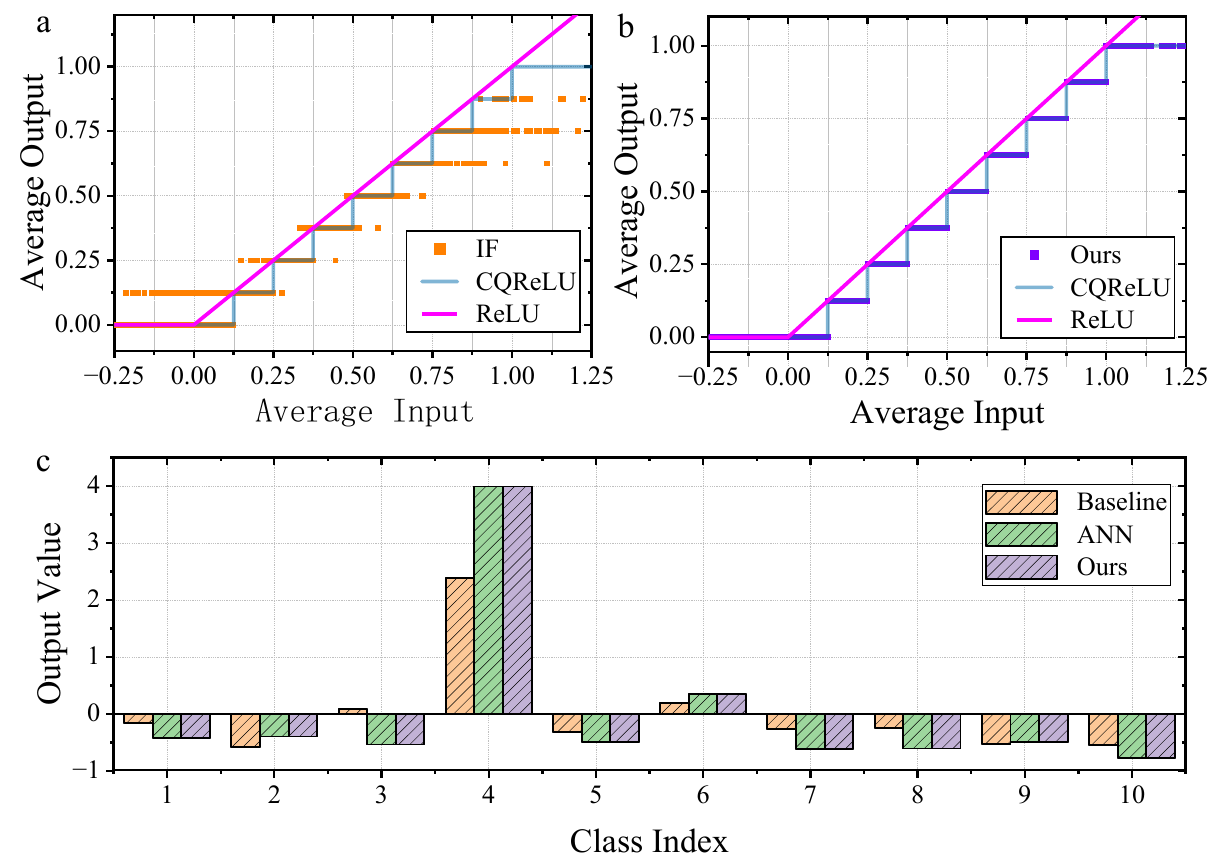}
	\caption{Statistics of the mapping of total input and output of spiking neurons and the output of SNN and ANN on CIFAR10 with VGG16. (a) Traditional methods using IF neurons (orange points) fail to achieve a one-to-one mapping relationship in total input-output as ANNs (pink line), (b) whereas our method (purple points) can achieve fitting with CQReLU (blue line). (c) Visualizing the outputs of the three models, we find that our method accurately identifies the maximum values and closely approximates the ANN outputs without loss.}
  \label{consistent}
\end{figure}

In extending the advantages of SNNs in terms of low energy consumption and fast inference, neither method can make SNNs perform as well as ANNs. Therefore, to thoroughly combine the advantages of backpropagation and SNN, ANN-SNN conversion methods are proposed \cite{diehl2015fast,han2020rmp}. However, they usually suffer from severe time delays and performance degradation. Yan et al. \cite{yan2021near} clarify that because of the discrete nature of spikes and the fact that the total firing rate does not exceed 1, SNNs cannot approximate arbitrary activation values of ANNs. Therefore, to impose restrictions on the ANN training process, as shown in Fig.\ref{consistent} (a) and (b), CQReLU, which has a stepped shape, is proposed. In addition, the problem of more and less spikes has been gaining attention. Li et al. \cite{li2022efficient} argue that the spikes in an SNN with zero activation value in the ANN seriously undermine the approximation principle and cannot be compensated by increasing the simulation time. Bu et al. \cite{bu2021optimal} think that this error, owing to the spike unevenness, makes the accurate approximation of the ANN difficult. The analysis of the conversion error has driven the progress of conversion methods. It enables SNN for target detection \cite{kim2020spiking}, semantic segmentation \cite{li2022spike}, and target tracking \cite{luo2022conversion} tasks. However, a unified framework must still explain the causes of conversion errors and guide the implementation of loss-free conversion. Although some SNNs show impressive performance in classification, they usually require a hundred steps to achieve the same performance as ANNs and rarely report the results on detection tasks.

To explain the source of the conversion error, we provide an in-depth derivation of the conversion process from the perspective of the mapping relationship between the total input and the output of the activation function. We deeply analyze the SNN output's approximation for different input cases. As shown in Fig.\ref{consistent} (a) and (b), the inconsistency in the mapping relationship between the ANN activation function and the total input and output of spiking neurons can be concluded to be the fundamental cause of conversion errors. As depicted through the pink line, the ANN activation function mapping is one-to-one, while the total input and output of IF neurons show a step-like, non-one-to-one mapping. Consequently, we utilize CQReLU in the ANN to induce a step-shaped activation function. By analyzing the necessary upper and lower limits for precise conversion, we propose a consistent ANN-SNN conversion (CASC) framework to achieve lossless conversion. It contains consistent IF neuron (CIF) and wake-sleep conversion (WSC) methods, as represented by the purple points in Fig.\ref{consistent} (b). This enables the output of the final layer of the SNN to not only produce the desired maximum values but also numerically match that of the ANN, as depicted in Fig.\ref{consistent} (c). Our contributions can be summarized as follows:

\begin{itemize}
	\item We provide an in-depth analysis of the conversion process from the perspective of the mapping relationship of activation function in ANN and SNN. We claim that the inconsistency of the mapping relationship of the ANN activation function and that between the input and output of the spiking neurons is the root cause of the conversion error.
	\item We analyze the upper and lower bounds that must be satisfied to achieve accurate conversion. Subsequently, we propose the CIF model and WSC, which can theoretically achieve lossless conversion with little delay.
	\item We perform extensive experiments on classification and detection tasks to verify the effectiveness of the proposed algorithms. Our approach outperforms the previous state-of-the-art methods and demonstrates considerable superiority in all datasets and structures.
\end{itemize}

\section{Related Work}

In recent years, training methods for obtaining deep spiking neural networks are mainly divided into two types.\\
\textbf{Direct training with surrogate gradient.} 
Surrogate gradients achieve direct training of SNNs using surrogate functions \cite{wu2018spatio,shrestha2018slayer,neftci2019surrogate} instead of delta functions as gradients. SpikeProp \cite{bohte2002error} first used smooth functions to overcome the non-differentiability problem of SNN. Wu et al. \cite{wu2018spatio} proposed the STBP method to optimize SNNs with BPTT, which has been widely employed in recent studies.
However, the gradient mismatch often makes the determination of the optimal solution for SNNs with a fixed form of SG challenging. Many researchers have attempted to optimize the performance of SNNs using gradual surrogate gradients \cite{guoloss,chen2022gradual}, replacing the Heaviside function with differentiable spikes \cite{li2021differentiable}, searching for optimal SGs and structures \cite{chedifferentiable}, and setting efficient loss functions \cite{deng2022temporal}. In addition, the lateral inhibition mechanisms \cite{cheng2020lisnn}, attention mechanisms \cite{yao2021temporal}, adaptive-parameter neurons \cite{fang2021incorporating}, parallel mechanisms \cite{li2024efficient}, and synaptic delay \cite{hammouamri2023learning} are effectively combined with SNNs.
Nonetheless, the abovementioned method performs worse on ImageNet and more complex tasks. They all require a large amount of memory and computational resources, limiting SNN application in complex scenarios.\\
\textbf{ANN-SNN Conversion. }
By mapping the parameters of a trained ANN to an SNN with the same topology, this method obtains high-performance deep SNNs. Cao et al. \cite{cao2015spiking} first applied the conversion method to train SNNs. Diehl et al. \cite{diehl2015fast} further enhanced the performance by normalizing the parameters. Rueckauer et al. \cite{rueckauer2017conversion} then achieved a more robust conversion by selecting smaller activation values for normalization. In addition, researchers proposed a soft reset mechanism \cite{rueckauer2017conversion,han2020rmp} to reduce information loss, which has been widely employed in subsequent studies. Recent work has attempted to achieve comparable performance with ANN at fewer time delays by dynamically adjusting the threshold \cite{han2020rmp,he2024msat}, correcting the weights \cite{li2022converting}, initializing the membrane potential \cite{bu2022optimized}, utilizing burst spikes \cite{li2022efficient}, adjusting the activation function of ANN \cite{yan2021near}, or using hybrid methods \cite{rathi2020enabling}.
Although these methods perform better in shorter time steps, none can achieve loss-free conversion. Therefore, applying them to more complex object detection tasks requires much work. Several attempts on detection tasks \cite{kim2020spiking,li2022spike,kim2020towards} required hundreds to thousands of time steps to achieve comparable results with ANN. {This paper aims to provide an accurate approximation of ANNs by SNNs, that is, the firing rate of neurons in SNNs is strictly equal to the quantized activation values of ANNs, thus achieving lossless conversion in fewer time steps in these tasks.}

\section{Preliminary}

Let the vector after convolution layer in ANN be $\textbf{z}^{(\ell)}$ and the vector after activation be $\textbf{a}^{(\ell)}=h(\textbf{z}^{(\ell)})$, where $h(x)=\operatorname{ReLU}(x)$ is the ReLU activation function. Therefore, the information transfer process in ANN can be expressed as
\begin{align}
	\textbf{a}^{(\ell+1)} = h(\textbf{z}^{(\ell+1)}) = h(\textbf{W}^{(\ell+1)}\textbf{a}^{(\ell)}),\quad 0<\ell \leq L
\end{align} where $\textbf{W}^{(\ell)}$ is the synaptic weight of $\ell$-th layer and $L$ is the total layers of the net. Converted SNNs typically use IF neurons, in which the membrane potential $\textbf{V}^{[\ell]}$ is updated after receiving spikes $\textbf{s}^{(\ell-1)}$ from a presynaptic neuron, and when the membrane potential exceeds a threshold $\textbf{V}_{th}^{(\ell)}$, the neuron delivers a spike. 
\begin{align}
	V^{(\ell)}_i[t] = V^{(\ell)}_i[t]+W_i^{(\ell)}s^{(\ell)}[t-1] \\
	s_i^{(\ell)} = \begin{cases}
		1, V^{(\ell)}\geq V_{th},\\
		0, else
	\end{cases}
\end{align}
To reduce the loss of information transfer process, we use a soft reset \cite{rueckauer2017conversion,han2020rmp}, i.e., $V^{(\ell)}[t] \leftarrow V^{(\ell)}[t]-s^{(\ell)}[t]$. 

Previous work considers that the firing rate of IF neurons can approximate the activation value of neurons in ANN. More effort is put into reducing the time delay and performance loss. Some works have detailed the approximate relationship between the IF neuron and ReLU activation function. Nonetheless, the key problem is whether the activation values of artificial neurons can be accurately mapped to the firing rates of spiking neurons. We assume that the total input of spiking neurons is $X^{(\ell)}_i[t]=\sum _{\tau=0}^t I^{(\ell)}_i[t]$ and the total output is $Y^{(\ell)}_i[t]=\sum _{\tau=0}^t s^{(\ell)}_i[t]$. Does the relationship between $Y^{(\ell)}_i[T]$ and $X^{(\ell)}_i[T]$ satisfy the same mapping rules as $z_i^{(\ell)}$ and $a_i^{(\ell)}$? where $T$ denotes the total time steps. Exploring them helps us to achieve a more accurate conversion.

\section{Methodology}
\subsection{Rethink the conversion process}

\textbf{Case 1:  Real value input.} To obtain better performance, converted SNNs are usually represented with real numbers in the input and output layers. Therefore, for the first layer, $\textbf{a}^{(0)}=\textbf{s}^{(0)}=\textbf{x}$, where $\textbf{x}$ is the input vector. For ANNs, $\textbf{a}^{(1)} = h(\textbf{z}^{(1)}) = h(\textbf{W}^{(1)} \textbf{a}^{(0)})$. And for SNNs, because $\textbf{I}^{(1)}[t]=\textbf{W}_i^{(1)}\textbf{s}^{(0)}[t]=\textbf{W}_i^{(1)}\textbf{x}$ is constant, the total input received increases monotonically with time steps if $a_i^{(1)}$ is greater than 0, i.e., $X_i^{(1)}[t+1] \geq X_i^{(1)}[t]$ and $\textbf{X}^{(1)}[T]=\sum\limits_{t=0}^{T}\textbf{I}^{(1)}[t]=T\textbf{W}_i^{(1)}\textbf{x}=T\textbf{z}^{(1)}$. Considering that the spikes of the IF neuron are cumulative discharges of the membrane potential, its total output can be expressed by the following equation.
\begin{align}
	\textbf{Y}^{(1)}[T] = \sum \limits _{t=0}^{T}\textbf{s}^{(1)}[t] = \operatorname{Clip}\left(\left\lfloor \frac{T\textbf{W}^{(1)}\textbf{
	}}{V_{th}} \right\rfloor, 0, T\right)
\end{align} Here, the $\operatorname{Clip}$ function defines the upper bound $T$ and lower bound 0 of the input; $\lfloor x \rfloor$ then returns the largest integer less than or equal to $x$. To simplify the operation, the threshold value of neurons is set to 1. Thus, we can get
\begin{align}
	\frac{\textbf{Y}^{(1)}[T]}{T} = \operatorname{Clip}\left( \frac{\left\lfloor \textbf{X}^{(1)}[T]\right\rfloor}{T}, 0, 1 \right)
	\label{ann}
\end{align}
In Case 1, the IF neuron performs additional upper-bound taking and quantization operations on the total input current when compared to the ReLU activation function. We call it a Clip-quantization (CQ) error. The most straightforward way is to modify the ReLU function in ANN to match the characteristics of the SNN and train the ANN using the CQReLU function with the following representation:
\begin{align}
	\operatorname{CQReLU}(x, \mathcal{Q}) = \operatorname{Clip}\left( \frac{\left\lfloor x\mathcal{Q}\right\rfloor}{\mathcal{Q}}, 0, 1 \right)
\end{align}
In this paper, we adopt CQReLU to train ANN. However, considering $x\in \mathbb{R}$, we can get the following problem:
\begin{align}
		\frac{Y^{(1)}[T]}{T} \geq \operatorname{CQReLU}(x, \mathcal{Q}), \quad when \quad T \geq \mathcal{Q}
		\label{eq7}
\end{align}
For example, if $x=0.58, T=16, \mathcal{Q}=8$, then $\frac{Y^{(1)}[T]}{T}=0.5625$ and $\operatorname{CQReLU}(x, \mathcal{Q})=0.5$. This shows that when $T \geq \mathcal{Q}$, the output of the SNN will also be larger than expected, i.e., the information output by the first layer of the network will no longer be accurate.

 \begin{figure*}[t]
	\centering
	\includegraphics[scale=0.48]{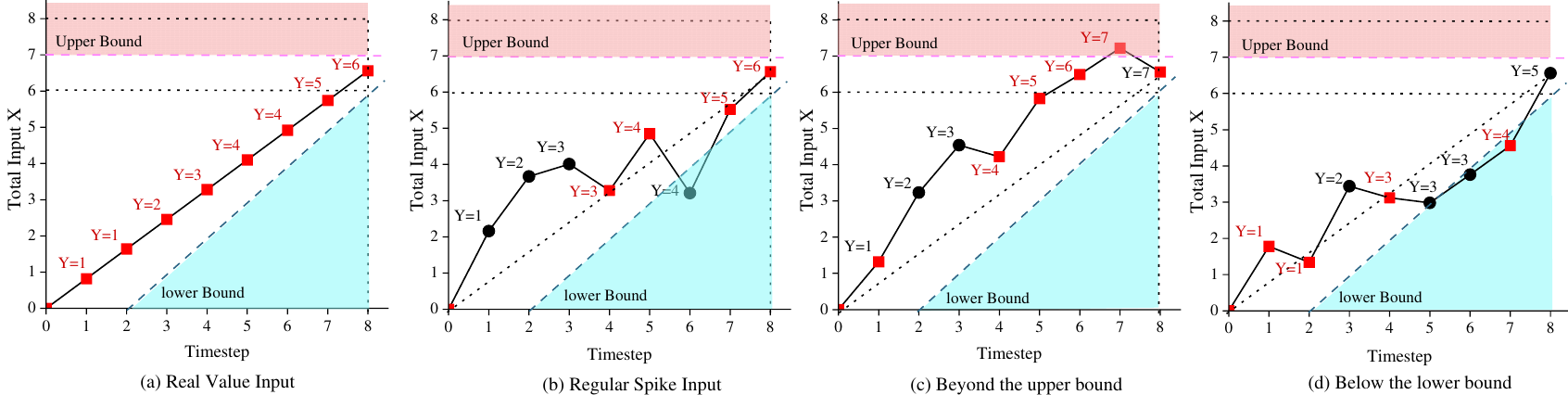}
	\caption{Impact of stable points on achieving consistent input-output mapping with ANN. We show how $Y$ varies with four different cases of inputs $X$, where the red squares are stable points. (a) The stable point does not cross the upper and lower bounds for real-valued inputs. (b) As long as the stable point does not cross the upper and lower bounds, the SNN can obtain the expected output. (c) If the stable point exceeds the upper bound, the SNN cannot recover the spikes already issued, thus causing more spikes. (d) If the stable point is below the lower bound, the SNN does not have the extra time step to emit the spikes that should have been transmitted.}
	\label{stab}
\end{figure*}

\textbf{Case 2:  The spike train's firing rate equals the ANN's activation value.} For discussion purposes, we assume that the activation values of the neurons in the ANN all satisfy Eq.\ref{ann}, i.e., no CQ error in the conversion process exists. We must focus on what relationship the total synaptic current in this layer satisfies with the total spikes issued when the firing rate of the IF neuron in the last layer is equal to the corresponding ANN activation value. We assume that the spiking activity of the IF neurons in the last layer follows a Poisson distribution, i.e., $Y^{(\ell-1)}_i[t] \sim Possion(a^{(\ell-1)}_i)$, For easy representation, we set $0 \leq a^{(\ell-1)}_i \leq 1$. For the total simulation time T, we have $Y^{(\ell-1)}_i[T] = Ta^{(\ell-1)}_i$. Thus, 
\begin{align}
	X_i^{(\ell)}[T] = \sum\limits_{t=0}^T \sum _j w^{(\ell)}_{ij}s_j^{(\ell-1)}[t] = Tz_i^{(\ell)}
\end{align}

Because of the event nature of SNNs, the synaptic current at each time step can be shown as the sum of the weights of the synapses connecting the spiking neurons.
Then, the probability that the current takes $w_{1j}^{(\ell)}$ can be represented by
\begin{align}
	\label{pro}
	P\left(I^{(\ell)}_i[t]=w_{ij}^{(\ell)}\right)=a_1^{(\ell)} (1-a_2^{(\ell)}) \cdots (1-a_n^{(\ell)})
\end{align}

%Since $w_{ij}^{(\ell)}$ obeys a Gaussian distribution,  so $Y^{(1)}[t] = \sum \limits _{\tau=0}^{t}s^{(1)}[\tau] $ is not necessarily equal to $\operatorname{Clip}\left(\left\lfloor {tW^{(1)}x} \right\rfloor, 0, t\right)$.
%
%Considering the characteristics of spiking process, i.e., $Y^{(1)}[t] = \Theta(X^{(l)}[t]-Y^{(l)}[t-1]-V_{th})$, where $\operatorname{geq}$ is used to determine the magnitude of the difference between the current total received and the historical total issued in relation to the threshold value,  Equation 5 does not necessarily hold. 
Thus, the synaptic current, i.e., the increment of $X_i^{(\ell)}[t]$, can be an arbitrary real value. $X_i^{(\ell)}$ has the possibility of arbitrary variation within $(0,T]$.
Owing to the unpredictability of synaptic currents and the 0/1 nature of the spikes, the value of $Y_i^{(\ell)}$ cannot simply depend on the instantaneous value of $X_i^{(\ell)}$.
Considering that it is not possible to obtain an exact expression for the variation of $Y_i^{(\ell)}$ with time steps, we can only explore the relationship between $X_i^{(\ell)}$ and $Y_i^{(\ell)}$ directly. Next, for ease of description, we use $X$ and $Y$ to denote $X_i^{(\ell)}$ and $Y_i^{(\ell)}$, respectively.
Define $\mathcal{T}= \{t|Y[t]=\left\lfloor X[t] \right\rfloor\}$ as the stable point set in the whole time steps, where $0\leq t_0<t_1<t_2<...<t_N\leq T$. For the convenience of representation, we assume that $\left\lfloor X[T] \right\rfloor \leq T$. Thus $Y[t_0]=\left\lfloor X[t_0] \right\rfloor = 0$.

Owing to the characteristics of spikes, the number of spikes delivered cannot exceed the simulation time, and the short-time stimulus reduction cannot change the spiking history. Thus, the relationship between X and Y can be expressed as
\begin{align}
	\label{eq10}
	Y[t_{n-1}] \leq  \left\lfloor X[t_n] \right\rfloor \leq \left\lfloor X[t_n] \right\rfloor +t_n - t_{n-1}
\end{align}The success of the conversion method shows that usually $T$ is the stable point, i.e., $Y[T]=\left\lfloor X[T] \right\rfloor$, as shown in Fig.\ref{stab} (a) and (b). From Eq.\ref{eq10}, we can conclude that accurate conversion requires two conditions: 

\textbf{Upper bound}:   $\forall t \in \mathcal{T}, Y[t]\leq \left\lfloor X[T] \right\rfloor$;
 
\textbf{Lower bound}: $\left\lfloor X[T] \right\rfloor- \left\lfloor X[t_{N-1}] \right\rfloor \leq T-t_{N-1}$.  

When larger synaptic currents tend to arrive in the early time steps, the neuron easily reaches the upper bound during the simulation, which results in more spikes, $V[T]\leq X[T]-\left\lfloor X[T] \right\rfloor$, as shown in Fig.\ref{stab} (c); conversely, the neuron does not have enough time to release the information in the membrane potential as spikes later in the simulation, $V[T]\geq X[T]-\left\lfloor X[T] \right\rfloor$, thus causing less spikes shown in Fig.\ref{stab} (d). Notably, non-stable points beyond the upper and lower bounds do not affect the accuracy of the output since Y has a delayed effect on X.

The derivation discussed above is based on the assumption that the input spikes satisfy a Poisson distribution. In particular, for spike trains with equal margins, the maximum convention of all spike intervals may be greater than or equal to the simulation time. The conversion error will also occur since Eq.\ref{pro} still holds and $I_i^{(\ell)}[t]$ is unstable and unpredictable. For instance, suppose the input to the network is real-valued. Subsequently, the second layer of the network receives a uniform spike train; however, there might also be conversion errors.

\textbf{Case 3: The spike train's firing rate does not equal the ANN's activation value.} In this case, $X_i^{(\ell)} = \sum\limits_{t=0}^T \sum _j w^{(\ell)}_{ij}s_j^{(\ell-1)}[t] \not= Tz_i^{(\ell)}$, If the mapping relationship from $X_i^{(\ell)}$ to $Y_i^{(\ell)}$ satisfies the $\operatorname{CQReLU}$, the firing rate of the neurons in that layer will not be equal to the activation value of the corresponding layer. Case 3 will always positively feed forward to affect the fitting of the subsequent layers. If the mapping relationship does not satisfy the  $\operatorname{CQReLU}$, the SNN approximates the ANN more than expected.
Exploring the conversion error in deeper layers is meaningless because, equivalently, we are converting a neural network that has changed starting from that layer.

{Above, we discuss the mapping of the input of the convolutional or fully concatenated layer to the output for different input cases. We conclude that the inconsistency of the SNN-ANN mapping relationship is the main reason for the conversion failure. The error is generated gradually from the first layer, which causes the subsequent layers to enter Case 3 one after another, leading to the conversion's performance loss. Moreover, most of the previous work was assumed the fact that Case 2 would occur at every layer of the network, resulting in the inability to fundamentally eliminate conversion errors. Therefore, to achieve accurate conversion, we must remove the influence of the upper and lower bounds in Case 2. Subsequently, we can achieve consistent conversion, where the mapping relationship between the total input and output of the SNN is the same as that of the ANN activation function.}

\subsection{Consistent IF Model}

 \begin{figure*}[t]
	\centering
	\includegraphics[scale=0.43]{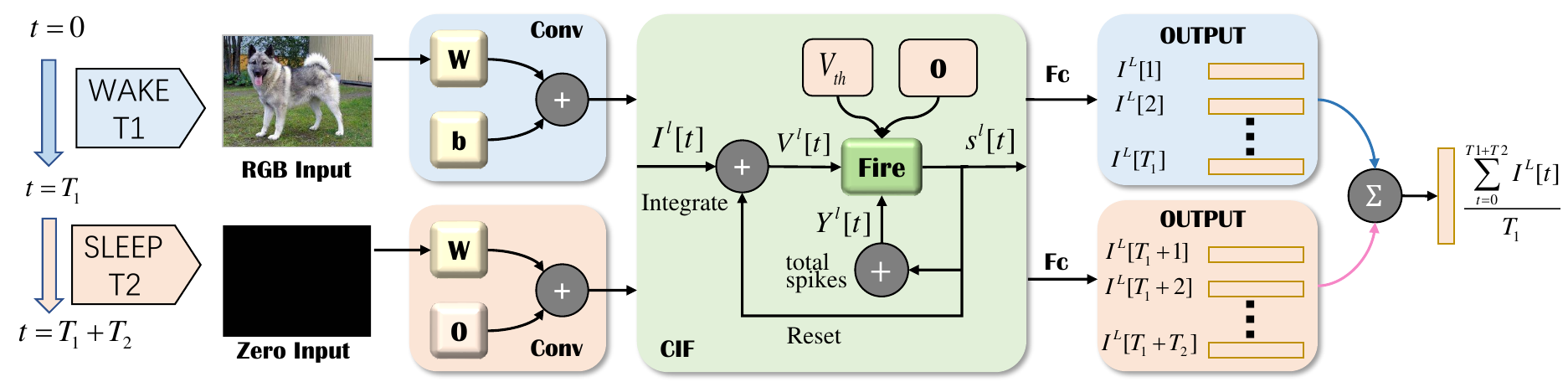}
	
	\caption{Illustration of our proposed method. In contrast to IF neurons, the CIF model also determines whether the current membrane potential is less than zero to ensure that neurons with a total historical output that is not zero satisfy $0\leq V_i^{(\ell)}[t]\leq V_{th}$, thus relieving the effect of the upper bound on accurate conversion. With WSC, the neurons only receive external information before $\mathcal{Q}$ time steps, ensuring the network has the same inputs as the ANN. Furthermore, WSC gives the SNN more time to emit unissued spikes and correct the more spikes. Using the proposed method, the SNN can accurately approximate the ANN activation value layer by layer.}
 \label{casc}
\end{figure*}

Ideally, we expect that all time steps during the simulation $T$ are stable points. In fact, it is sufficient that all stable points do not exceed the upper and lower bounds that satisfy the consistent conversion. We propose consistent IF (CIF) neurons to overcome the effect of the upper bound on the conversion, as shown in Fig. \ref{casc}. Since information is transmitted layer by layer in the SNN, cannot wait for the end of the simulation time to precisely correct for more spikes before passing the information to the next layer. Therefore, we think that the neuron must only believe that the total input up to the current time step has satisfied Case 2, i.e., $X_i^{(\ell)}[t] = tz_i^{(\ell)}$. To ensure a consistent mapping of inputs to outputs as ANN, the neuron judges whether to emit a spike based on the total input $X_i^{(\ell)}[t]$ and the total historical output $Y_i^{(\ell)}[t-1]$. The neuron emits spikes when it exceeds the threshold $V_{th}$. If the difference is less than zero and the total historical output is greater than zero, the neuron outputs a negative spike to pull the neuron back from the upper boundary. Thus, CIF can be formulated as the following:
\begin{align}
	s_i^{(\ell)}[t] &= \begin{cases}
		1, &V_i^{(\ell)}[t]\geq V_{th},\\
		-1, &V_i^{(\ell)}[t] < 0 \quad and \quad Y_i^{(\ell)}[t-1] > 0\\
		0, &otherwise
	\end{cases}
\end{align}

Here, $Y_i^{(\ell)}[t]=Y_i^{(\ell)}[t-1]+s_i^{(\ell)}[t]$ is the total output. The CIF neuron must only memorize the total output spikes additionally and not the total inputs because $V_i^{(\ell)}[t]=X_i^{(\ell)}[t]-Y_i^{\ell}[t-1]$. 
The SNM \cite{wang2022signed} has a similar form to the CIF, but the SNM only issues a negative spike when the membrane potential is less than a negative threshold, so a portion of the neurons that exceed the previous session is not adjusted. By contrast, the goal of CIF is that the neuronal membrane potential should be between zero and the threshold when the total spike output is not zero.

\subsection{Wake-Sleep Conversion}

The CIF model adjusts the neuron automatically when the upper bound is exceeded. However, its magnitude is limited and may not be fully adjusted in subsequent time steps. In addition, the number of spikes is limited by the difference between the simulation steps and the current time step. Therefore, to reduce this effect, previous studies aim to increase the simulation step size. However, from the perspective of the total number of errors, the network also receives new inputs in the increased simulation step, and this part of the error is re-enacted. Therefore, we propose a wake-sleep conversion (WSC) framework in Fig. \ref{casc}. In the wake phase, the model receives the RGB input and passes the information layer by layer, as in the previous conversion method. {Then, after $\mathcal{Q}$ time steps, the conversion process enters the sleep phase. The model does not receive new external inputs but only transfers information within the model. Subsequently, regardless of the total simulation steps, the model only accepts external information for $\mathcal{Q}$ time steps. The bias of the convolutional or fully connected layers is considered an additional input current to the neuron. During the sleep phase, we set the bias in all the layers to zero so that the SNN accepts only $\mathcal{Q}$ time steps of extra current from the bias and focuses only on the spikes information delivered by the front layer during the sleep phase. The neuron then has the opportunity to further release the incompletely released information and adjust the incompletely adjusted spikes. After the adjustment in the sleep stage, the consistency condition of Case 2 is satisfied layer by layer from the shallow layer to the deep layer. Thus, the output information of SNN can completely approximate the output of ANN when the simulation time is sufficient. }
\begin{align}
	\frac{Y^{(1)}[T]}{T} = \operatorname{CQReLU}(x, \mathcal{Q}), \quad when \quad T \geq \mathcal{Q}
\end{align}

\section{Experiments}
In this section, we validate the effectiveness and efficiency of the proposed consistent conversion method while visualizing the error of the conversion process. 
The experimental results show that our method considerably outperforms existing SOTA methods in terms of performance and inference speed. 

\paragraph{Classification Task.} All the experiments are carried out on the NVIDIA A100 with the PyTorch framework. We first test on the CIFAR10 and CIFAR100 datasets with VGG16 and ResNet20 architecture and then validate our model on the larger ImageNet dataset with VGG16 and ResNet34. $\operatorname{CQReLU}$ replaces the original ReLU activation function, and then we train the network directly. The total epochs are set to 300, batch\_size to 128, and the Cosine learning rate decay strategy is employed. For the CIFAR dataset, we use the AdamW optimizer, and for ImageNet, we use the SGD optimizer. The initial learning rate for both is 0.1, and the weight decay is 1e-4. To enhance the performance of the ANN model, the label smoothing technique is used with a factor of 0.1. We use Cutout \cite{devries2017improved} and Autoaugment \cite{cubuk2019autoaugment} techniques for both datasets. As in the previous work \cite{li2021bsnn,li2021free,li2022converting,kim2020spiking}, during SNN inference, we use real values in both the input and output layers.

\paragraph{Detection Task}
We test the object detection tasks on PASCAL VOC and COCO datasets with a modified YOLO the same as \cite{li2022spike}. It employs VGG16 as the backbone for feature extraction, followed by SPP \cite{he2015spatial} and a convolutional module as the neck. 
A convolutional module comprises a convolutional layer, batch normalization layer, and activation function layer. Subsequently, a classification head consisting of two convolutional modules and a regression head comprising four convolutional modules are followed parallelly. The last three parallel convolutional layers are used to judge the object, classification, and regression. For the VOC dataset, we use 16,552 samples from the training set of VOC2007 and the validation set from VOC2012 for training. We use 4,952 test samples from the test set of VOC2007 as the test set. For the COCO dataset, we use the original 118,287 training samples for training and 5,000 test samples for testing. For comparison with previous work, we use the mAP@0.5 metric for the VOC dataset and the mAP@0.5:0.95 metric for the COCO dataset. To improve the performance of the model, we use multi-scale training \cite{singh2018sniper}, and warm-up \cite{he2016deep} methods.

\subsection{Influence of Quantization Level}
\begin{table}[ht]
	\centering
	\resizebox{0.7\columnwidth}{!}{    %width, columnwidth
		\begin{tabular}{l|l|ccccc}
			\toprule
			Network&Accuracy & $\mathcal{Q}$=None & $\mathcal{Q}$=64 & $\mathcal{Q}$=32 &$ \mathcal{Q}$=16 & $\mathcal{Q}$=8\\
			\midrule
			\multirow{3}{*}{\makecell{VGG16,\\CIFAR10}}
			&ANN & 95.74 & 95.82 & 95.71& 95.14 & 94.62\\
			&SNN Best & 95.55 & 95.84 & 95.78& 95.18 & 94.64\\
			&TimeStep & 33& 72 & 34& 19 & 9\\
			\midrule
			\multirow{3}{*}{\makecell{ResNet20, \\CIFAR100}}
			&ANN & 79.35 & 77.98 & 77.53& 78.30 & 77.75\\
			&SNN Best & 77.35 & 78.40 & 77.76& 78.30 & 78.17\\
			&TimeStep & 42 & 55 & 37& 21 & 12\\
			\midrule
			\multirow{3}{*}{\makecell{ResNet34, \\ImageNet}}
			&ANN & 75.46  & 72.83 &  72.71 & 72.20& 71.08\\
			&SNN Best & 59.80 & 72.89 & 72.89& 72.16& 71.05\\
			&TimeStep & 71 & 88 & 55 & 38& 24\\
			\bottomrule
	\end{tabular}}
	\caption{Conversion accuracy across different quantization levels with VGG16, ResNet20, and ResNet34 on CIFAR10, CIFAR100, and ImageNet Datasets. We demonstrate the robustness of our proposed method to quantization levels, showing the performance and time steps for the converted SNNs under the influence of different quantization levels.}
	\label{ab2}
\end{table}

Our experiments in the paper report results for the classification task at quantization levels $\mathcal{Q}=8$ and 16. To show that our proposed method is robust to $\mathcal{Q}$, we test the best performance and the latency of the converted SNN at different  $\mathcal{Q}$. We conduct experiments using VGG16, ResNet20, and ResNet34 on CIFAR10, CIFAR100, and ImageNet. The results shown in Tab.\ref{ab2}  demonstrate that our proposed method achieves the best performance within twice the simulation time of $\mathcal{Q}$ in all cases using $\operatorname{CQReLU}$. The conversion results in the case of using ReLU exhibit some performance loss. It demonstrates the correctness of our theoretical analysis of conversion errors in the paper, that is, achieving loss-free conversion requires restrictions on ANNs. The effectiveness of the proposed method is shown in Tab.\ref{ab2}.

\subsection{Effect of the Proposed Methods}

\begin{figure}[t]
	\centering
	\includegraphics[scale=0.6]{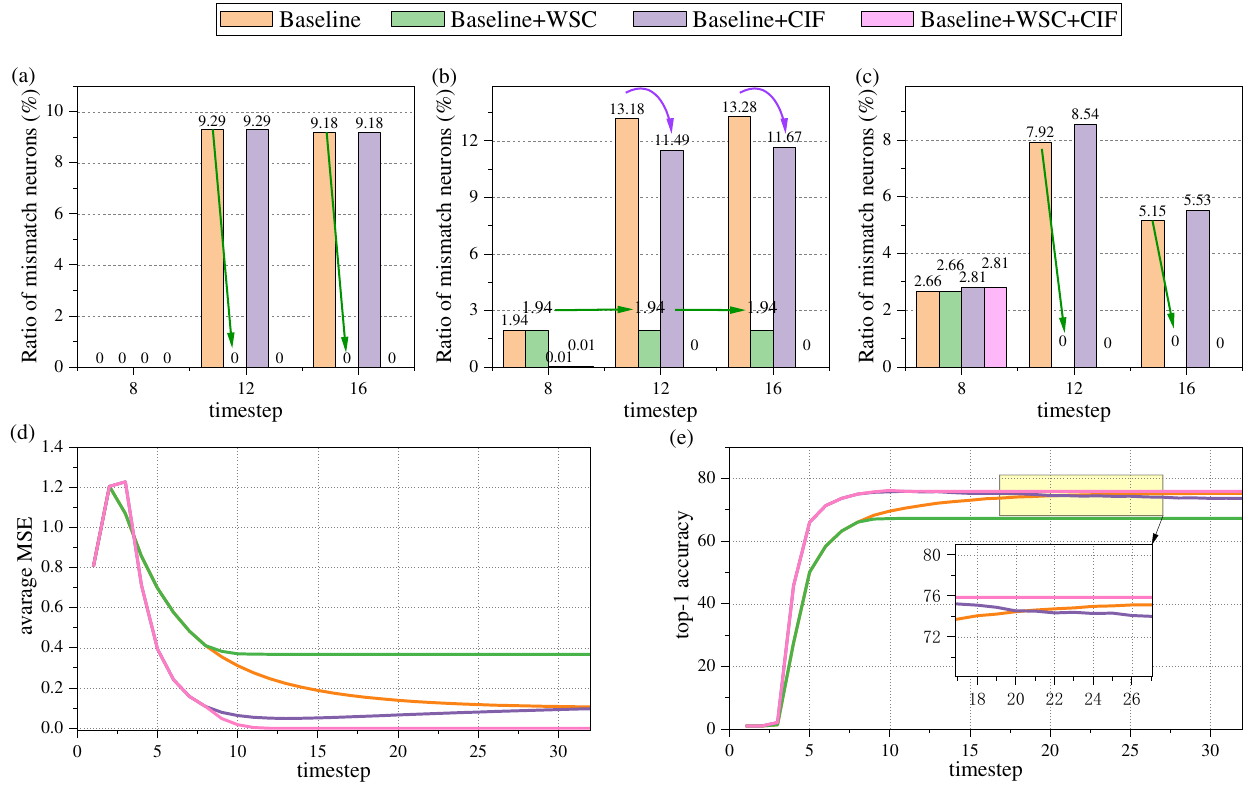}
	
	\caption{Effect of the proposed methods on CIFAR100 with VGG16. (a) Ratio of neurons with more spikes in the first layer. (b) Ratio of neurons with more spikes in the second layer. (c) Ratio of neurons with fewer spikes in the second layer. (d) MSE of SNN output and ANN output. (e) Top-1 accuracy of our methods.}
	\label{alab}
\end{figure}

To further explain the effect of WSC and CIF neurons, we visualize the mismatched ANN activation values ratio in the first and second layers of the network for VGG16 in the CIFAR100 dataset, where $\mathcal{Q}=8$. Fig.\ref{alab} (a) shows how the ratio of neurons with more spikes in the first layer of neurons varies at time steps 8, 12, and 16. Since the experiment uses real-valued inputs, there are no less spikes for the first layer neurons according to Eq.\ref{eq7}. 
The results reveal that WSC can solve the problems caused by using $\operatorname{CQReLU}$ mentioned in case 1.
Fig.\ref{alab} (b) and (c) show the ratio of neurons with more spikes and less spikes in the second layer, respectively. The green bar in Fig.\ref{alab} (c) indicates that the neuron may deviate from the expected value even when receiving a uniform spike train, corroborating our discussion in case 2. Another conclusion is that WSC can avoid the problem of more spikes in the first layer and solve the problem of less spikes in deep-layer neurons. In Fig.\ref{alab} (b), CIF can considerably reduce the problem of more spikes in the deep layer. It is not entirely solved, as shown in the purple bar, because some neurons in the first layer are mismatched without the help of WSC.

Fig.\ref{alab} (d) shows the MSE curves of the SNN and ANN outputs. As shown, the purple line is dynamically changing, while the green line quickly tends to stabilize. When only WSC is used, the output of the first layer of spike neurons is tuned to be consistent with ANN, and the network gradually passes the zero information from the shallow layer to the deep layer. However, the information about more spikes cannot be tuned. Finally, the network output tends to be stable and maintains some error with ANN.
In contrast, information from more spikes passed by the first layer cannot be processed when only CIF is employed. The increase in time steps partially resolves the ratio of neurons with less spikes in each layer. Moreover, neurons with more spikes are dynamically tuned by CIF, thus its output is also unstable. 
The classification accuracy curve of SNN shown in Fig.\ref{alab} (e)  exhibits the same trend as Fig.\ref{alab} (d). When both techniques are used, the activation value of the SNN matches the ANN, and the MSE of the network output converges to zero. Thus, exact conversion can be achieved.

\begin{figure}[!t]
	\centering
	\includegraphics[scale=0.475]{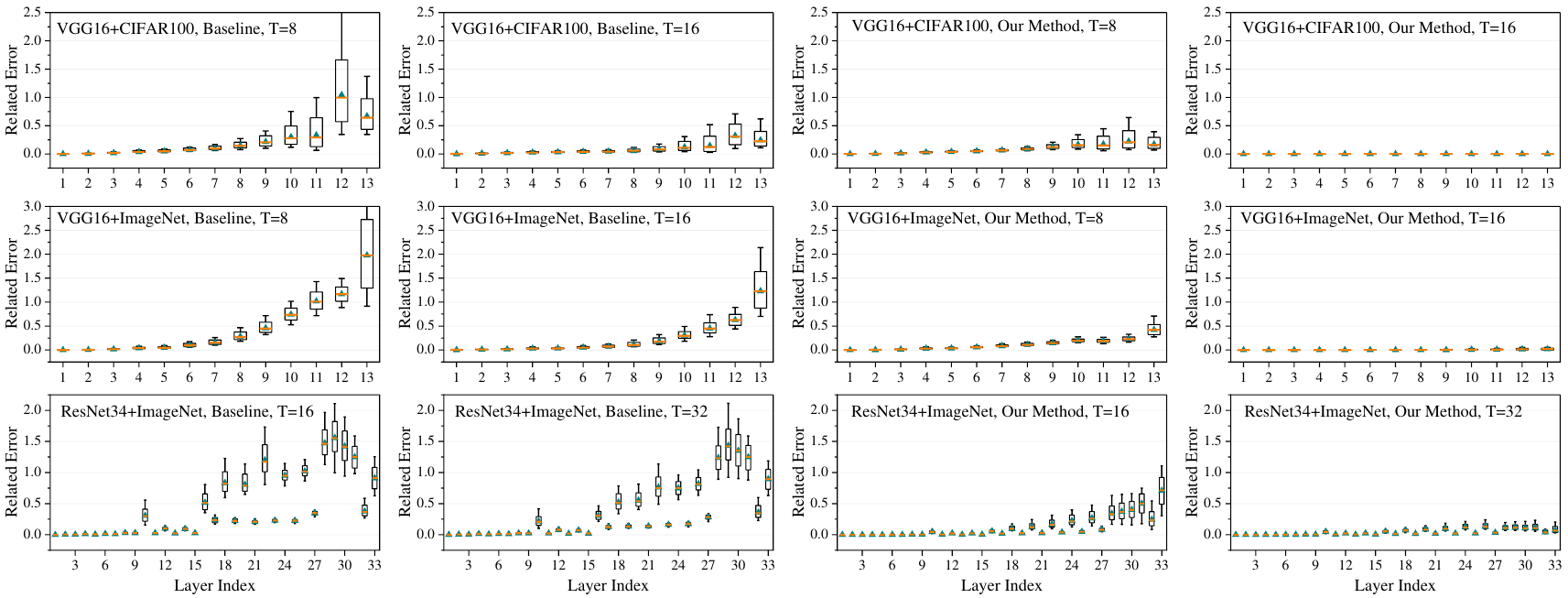}
	\caption{Measuring the relative error on the CIFAR100 and ImageNet with VGG16 and ResNet34. Our method demonstrates the ability to achieve nearly loss-free conversion. The horizontal line represents the mean, the triangle represents the median, and the box range is 5\% to 95\%.}
 \label{vis}
\end{figure}

\paragraph*{Visualization of Relative Error}
To further illustrate the effectiveness of the proposed method for achieving accurate conversion, we follow \cite{li2022converting} to visualize the relative error per layer on different networks and datasets, as shown in Fig.\ref{vis}. We randomly selected 1000 samples for statistics and set the quantification level to 8. In the VGG structure, WSC helps the CIF neuron handle the error at each layer, thus achieving a near-zero error conversion. The results on ResNet and ImageNet also show the accurate approximation to the ANN activation value. Given that more time is still needed, some layers do not achieve zero error.

 \begin{figure}[ht]
	\centering
	\includegraphics[scale=0.385
	]{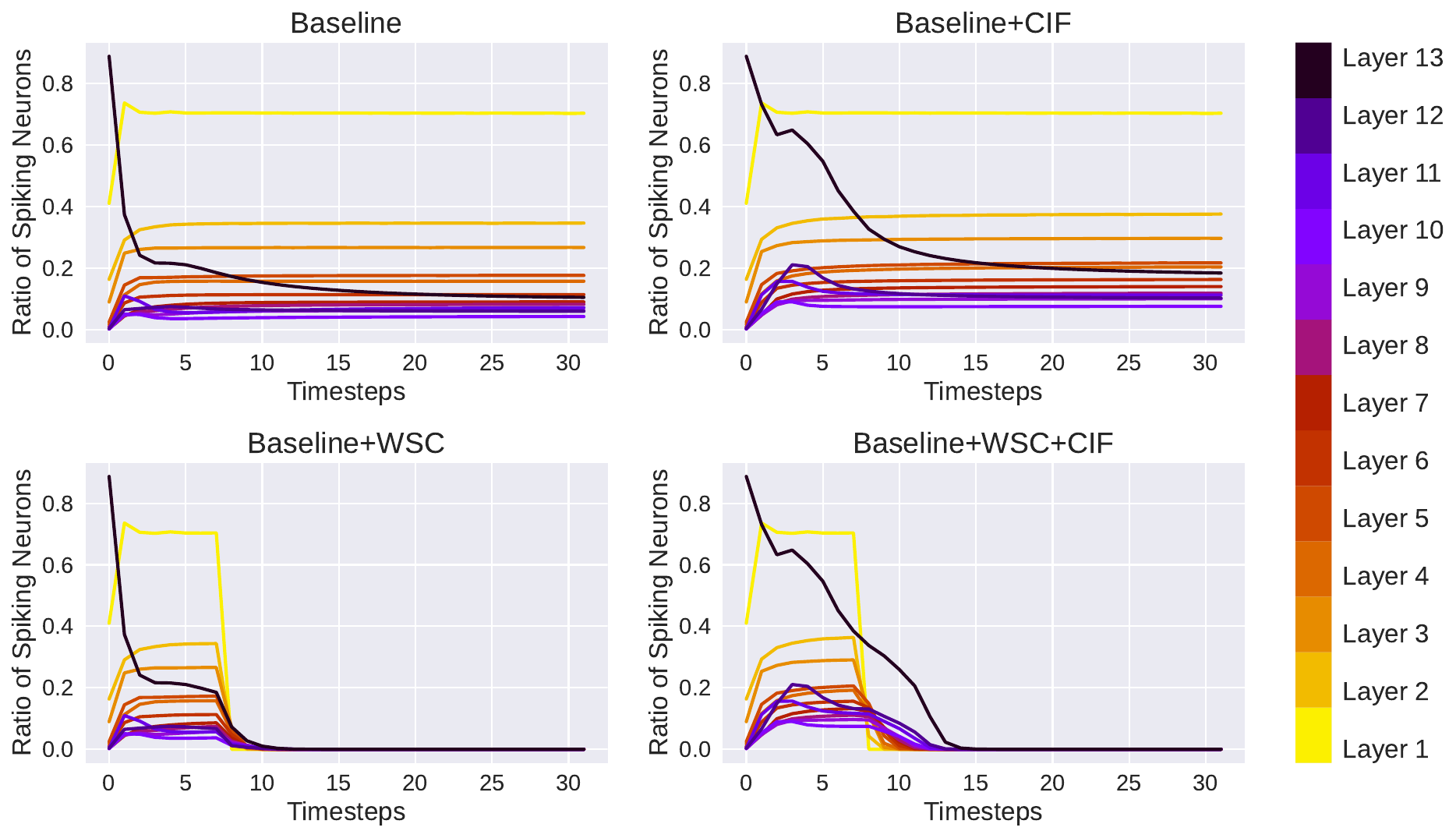}
	
	\caption{Curve of the ratio of firing neurons with the time steps. We perform statistics on the entire Imagenet dataset with VGG16 architecture. Our method goes to a resting state after the precise conversion, further reducing the energy consumption caused by excessive spikes, while traditional methods emit spikes continuously.}
	\label{spar}
\end{figure}

\paragraph*{Sparsity and Computational Efficiency}
The computational efficiency of the SNN is closely related to the spiking activity. We visualize in Fig.\ref{spar} the proportion of neurons emitting spikes for each layer of the VGG16 over the entire ImageNet dataset with time steps. We treat both positive and negative spikes as one. Without the proposed method, the spike activity of neurons in each layer gradually stabilizes over time, that is, it maintains a relatively stable firing rate. It leads to a positive correlation between the computational energy consumption and the inference time of the SNN. 
Moreover, in Fig.\ref{spar}, due to the WSC method, the ratio of the neurons emitting spikes is not considerably increased despite negative spikes. Furthermore, the neurons will no longer send spikes after specific time steps, considerably enhancing the sparsity of the spiking activity. The curve change shows that the network can achieve the correction layer by layer until the neurons in all layers no longer spike. In addition, WSC allows the SNN not to have to process real-valued inputs after the time step $\mathcal{Q}$, which involves many multiply-accumulate (MAC) operations rather than accumulate (AC) operation \cite{rathi2021diet}, thus remarkably improving computational efficiency.

Notably, in Fig.\ref{spar}, in the baseline + WSC model, each layer almost ceases to emit spikes simultaneously, while in the CASC model, there is a gradual cessation of spike emission from shallow to deep layers. This is due to the limited activity of IF neurons during the sleep phase without inputs, where they can only release residual membrane potentials in the form of a few spikes. In contrast, CIF neurons exhibit more dynamic characteristics, capable of emitting both positive and negative spikes. This facilitates the gradual adjustment of neurons from shallow to deep layers to align with the mapping of CQReLU, thereby achieving the precise approximation of the output layer, as shown in Fig. \ref{vis}.

\subsection{Comparison to Previous Work}

\begin{table*}[t]
	\centering
		\resizebox{\textwidth}{!}{    %\textwidth, columnwidth
		\begin{tabular}{c|cccccc|cccccc}
			\toprule
			\multirow{2}{*}{Method } & \multicolumn{6}{c}{VGG16} & \multicolumn{6}{c}{ResNet20}\\
			\cmidrule{2-13}
			& ANN  & T=4 &T=8 & T=16  & T=32 & T=64 
			& ANN & T=4 & T=8 & T=16 & T=32 & T=64\\
			\midrule
			SNM \cite{wang2022signed} & 74.13 & - & - & -& 71.80 & 73.69 
			& N/A & - & - & - &-&-\\
			OPI \cite{bu2022optimized} & 76.31 & - & 60.49 &70.72 & 74.82 & 75.97  
			& 70.43 &- & 23.09 & 52.34&67.18 & 69.96\\ 
			Burst Spikes \cite{li2022efficient} & 78.49 & -& -& -&74.98 & 78.26
			& 80.69 & - & - & - &76.39&79.83\\

			QCFS \cite{bu2021optimal} & 76.28 & 69.62 & 73.96 & 76.24 & 77.01 & 77.10 
			& 69.94 & 34.14 & 55.37 & 67.33  & 69.82  & 70.49\\
			ACP \cite{li2022converting} & 77.93 & 55.60 & 64.13 & 72.23 & 75.53 & -
			& 81.50 & 54.96 & 71.86 & 78.13 & 80.56 & -\\
			SRP \cite{hao2023reducing} & 76.28 & - & 75.42 & $<$76.42 & $<$76.45 & $<$76.37 
			& 69.94 & - & 59.34 & $<$64.71 & $<$65.50 & $<$65.82 \\
			
			\midrule
			Baseline ($\mathcal{Q}$=8) &75.85  &  27.51 & 66.03 & 73.49& 75.19 & 74.33 
			&  77.75 & 1.18 & 23.67 & 65.20 &73.91 & 73.20 \\
			Ours ($\mathcal{Q}$=8) & 75.85 & 45.74 & {74.96} & 75.85 & 75.85 & 75.85
			& 77.75  & 8.44 & 68.27 &77.76 & 77.76 & 77.76\\
			%			\midrule
			Baseline ($\mathcal{Q}$=16) & 77.58 & 1.99 & 48.07 & 72.82 & 77.00 & 77.11
			& 78.30  & 1.04 & 2.93 &46.66 & 73.12 &76.64\\
			Ours ($\mathcal{Q}$=16) & 77.58  &1.00 & 72.42 & {77.73} & {77.56} & {77.56}
			& 78.30  & 0.97 & 38.85 & 76.22 & 78.27  & 78.27\\
			\bottomrule
		\end{tabular}}
		\caption{Detailed comparison of conversion accuracy between our proposed SNN conversion algorithm and existing methods on the CIFAR100 dataset. We present a comprehensive evaluation using VGG16 and ResNet34 architectures, showcasing the performance of our algorithm at various time steps (T = 4, 8, 16, 32, 64) against other benchmark SNN methods.}
		\label{tab2}
	\end{table*}

In this study, to ascertain the efficacy of our proposed approach, we juxtapose our methodology against other state-of-the-art ANN-to-SNN conversion algorithms. The SNM \cite{wang2022signed} employs a similar negative spikes mechanism. Burst Spikes \cite{li2022efficient} and QCFS \cite{bu2021optimal} enhance performance and reduce latency through multi-spikes mechanisms and quantization techniques, respectively. In contrast, the ACP \cite{li2022converting} method, which uniquely corrects weights without preprocessing the ANN, represents a quintessential approach.
The SRP method \cite{hao2023reducing} is bifurcated into two distinct stages. For an equitable comparison, we aggregate the temporal durations of both stages, denoted as $t+\tau$.

Tab.\ref{tab2} shows the performance of our proposed method on CIFAR100. For accurate conversions, we achieve lossless conversions in both VGG16 and ResNet20 structures and can achieve top-1 accuracies of 75.85 and 77.76 ($\mathcal{Q}=8$) in both structures with few time delays, outperforming all other methods. Note that QCFS outperforms the original ANN at 64-time steps due to its error-prone approximation of the ANN and the fact that CIFAR100 is a relatively easy dataset.

\begin{table*}[!t]
	\centering
		\resizebox{\textwidth}{!}{    %\textwidth, columnwidth
		\begin{tabular}{c|ccccc|ccccc}
			\toprule
			\multirow{2}{*}{Method } & \multicolumn{5}{c}{VGG16} & \multicolumn{5}{c}{ResNet34}\\
			\cmidrule{2-11}
			& ANN  & T=8 & T=16  & T=32 & T=64 
			& ANN & T=8 & T=16 & T=32 & T=64\\
			\midrule
			OPT \cite{deng2021optimal} & 75.36 & - & - & 0.11 & 0.12 
			& 75.66 & - & - & 0.09 & 0.12\\
			OPI \cite{bu2022optimized} & 74.85 & 6.25 & 36.02 &64.70 & 72.47  
			& N/A &- & - & -&-\\ 
			SNM \cite{wang2022signed} & 73.18 & - & - & 64.78 & 71.50 
			& N/A & - & - & - &-\\
			Burst Spikes \cite{li2022efficient} & 74.27 & -& -& 70.61 & 73.32
			& N/A & - & - & - &-\\
			%			Dual-phase \cite{wang2022towards} & 74.88 & 62.51 & 70.13 & 73.44 & 74.68 
			%			& 73.43 & 61.20 & 67.77 & 71.66  & 72.65\\
			QCFS \cite{bu2021optimal} & 74.29 & - & 50.97 & 68.47 & 72.85 
			& 73.43 & 61.20 & 67.77 & 71.66   & 72.65\\
			ACP \cite{li2022converting} & 75.36 & - & 65.02 & 69.04 & 72.52
			& 75.66 & - & 51.67 & 64.65 & 71.30\\
			SRP \cite{hao2023reducing} & 74.29 & - & 61.37 & $<$69.35 & $<$69.43 
			& 74.32 & - & 67.62 & $<$68.40 & $<$68.61 \\
			\midrule
			Baseline ($\mathcal{Q}$=8) &73.31  &  0.61 & 4.91 & 35.95 & 58.33 
			&  71.08 & 0.11 & 0.16 & 0.23 & 0.50 \\
			Ours ($\mathcal{Q}$=8) & 73.31 & {66.09} & {73.29} & {73.29} & 73.29
			& 71.08  & 0.11 & 12.11 &71.05 & 70.94\\
%			\midrule
			Baseline ($\mathcal{Q}$=16) & 73.88 & 0.10 & 0.47 & 2.44 & 24.15
			& 72.20  & 0.11 & 0.16&0.81 &6.83\\
			Ours ($\mathcal{Q}$=16) & 73.88  &47.66 & 73.07 & 73.93 & 73.93
			& 72.20  & 0.22 & 2.62 & 71.41 & 72.11\\
			\bottomrule
		\end{tabular}}
		\caption{Detailed comparison of conversion accuracy between our proposed SNN conversion algorithm and existing methods on the large-scale ImageNet dataset. We present a comprehensive evaluation using VGG16 and ResNet34 architectures, showcasing the performance of our algorithm at various time steps (T = 4, 8, 16, 32, 64) against other benchmark SNN methods.}
		\label{tab3}
	\end{table*}

To validate the performance of our method on large-scale datasets, we test the performance on the ImageNet dataset and report the results in Tab.\ref{tab3}. QCFS shows some performance loss in both VGG16 and ResNet34 architecture. On large datasets, our method is superior. It performs beyond all state-of-the-art methods, exhibiting low latency and lossless conversion advantages. For the VGG16 structure, we achieve an accuracy of 73.29 with only 16 time steps, and for the ResNet34, an accurate conversion is achieved using only 32 time steps. 

\begin{table}[t]
	\centering
	\resizebox{0.8\textwidth}{!}{    %\textwidth, columnwidth
		\begin{tabular}{c|c|ccccc}
			\toprule
			Method  &  Model & ANN  & T=8 & T=16 & T=32 & T=64\\
			\midrule
			\multicolumn{7}{c}{PASCAL VOC}\\
			\midrule
			Spiking-YOLO \cite{kim2020spiking} & Tiny-YOLO & 53.01 & - & - & - & $<$51.83\\
			Two-Phase \cite{kim2020towards} & Tiny-TOLO & 53.01 & - & - & - & $<$46.66\\
			SpikeCalibration \cite{li2022spike} & YOLO & 67.48 & - & - & - & 63.43\\
			\midrule
			Baseline ($\mathcal{Q}$=4) &\multirow{6}{*}{YOLO} & 65.27  & 9.24 & 39.32 & 49.10 & 46.80\\
			Ours ($\mathcal{Q}$=4) & & 65.27 & 63.15 & 65.42 & 65.42 & 65.42 \\
			Baseline ($\mathcal{Q}$=8) & & 72.92 & 0& 16.75 & 48.83 & 62.44\\
			Ours ($\mathcal{Q}$=8) & & 72.92 & 24.96 & 73.03 & 72.94 & 72.94  \\
			Baseline ($\mathcal{Q}$=16) &    & 74.25  & 0 &  7.07 & 41.29 & 65.48 \\
			Ours ($\mathcal{Q}$=16) &   & 74.25 & 19.94 & 59.59 & 74.29 &74.28 \\
			\midrule
			\multicolumn{7}{c}{MS COCO}\\
			\midrule
			ACP \cite{li2022converting} & RetinaNet & 35.60 & - & - & 28.40 & 32.30\\
			ACP \cite{li2022converting} & Faster R-CNN & 37.00 & - & - & 31.00 & 34.20\\
			\midrule
			Baseline ($\mathcal{Q}$=4) & \multirow{4}{*}{YOLO} & 26.39  & 1.13 & 6.59 & 13.82 & 15.83 \\
			Ours ($\mathcal{Q}$=4) & & 26.39  & 26.14 &26.35 & 26.35 & 26.35 \\
			Baseline ($\mathcal{Q}$=8) &  & 29.24 & 0.26 & 6.41 & 17.20 & 23.09\\
			Ours ($\mathcal{Q}$=8) & & 29.24 & 7.36 & 29.21 & 29.24 & 29.24  \\
%			Baseline (Q\mathcal{Q}=16) &  & 29.77 & 0 &0.19 & 4.71 & 15.85 \\
%			Ours (Q\mathcal{Q}=16) &   & 29.77 & 3.73 & 15.02 & 29.87 & 29.88 \\
			\bottomrule
	\end{tabular}}
	\caption{Detailed comparison of conversion performance between our proposed method and existing methods in object detection tasks on VOC and COCO datasets. It demonstrates our method's capability to generalize beyond classification tasks, addressing the challenges of high time delay and performance degradation in object detection. We use  mAP@0.5 in VOC and mAP@0.5:0.95 in COCO as the evaluation metrics.}
	\label{det}
\end{table}

We further test whether our method can be generalized to the object detection task. Although the above methods perform well on classification tasks, they have yet to be reported on the object detection task. The few previous attempts either needed additional work to reduce the time delay or could not overcome the performance degradation. Specifically, Spiking-YOLO \cite{kim2020spiking} initially demonstrated the capabilities of SNNs in object detection tasks. Conversely, its team further minimized the temporal latency incurred during the conversion process through a two-stage methodology \cite{kim2020towards}, yet this approach still required several thousand-time steps. On the other hand, SpikeCalibration \cite{li2022spike} achieved performance comparable to that of ANNs with just 128-time steps, although at the expense of a considerably increased computational cost. The ACP method \cite{li2022converting} further reduced temporal delays but has yet to achieve an optimal balance between low latency and high performance.
We report the performance of our method using YOLO structures on two large-scale object detection datasets, VOC and COCO, as shown in Tab.\ref{det}.For a fair comparison, we use the evaluation metrics of mAP@0.5 and map@0.5:0.95 for VOC and COCO, respectively. The results show that our method can achieve the object detection task losslessly using only 32 time steps. The performance at 8 and 16 time steps also demonstrates the inference speed of the proposed method. We achieve 73.03 mAP with 16 time steps on VOC and 29.21 mAP performance on COCO. All results show that our method can approximate the ANN output exactly, thus outperforming previous conversion methods.

\subsection{Performance on Other Models}

To ensure a fair comparison with previous methods, our experiments above utilize the classic VGG and ResNet architectures. This raises the question of whether the CASC framework is compatible with more novel network structures. We perform experimental validations on RegNetX\_8GF \cite{radosavovic2020designing} and ResNeXt50\_32x4d \cite{xie2017aggregated} structures to address this. RegNetX\_8GF, known for its computational efficiency and optimized performance, utilizes a systematic approach to network design, offering an alternative to traditional architectures. In contrast, ResNeXt50\_32x4d introduces a modularized grouping method, using 32 groups of four convolutions each, to enhance feature representation while maintaining computational efficiency. 
Tab.\ref{tab33} shows that the experimental results exhibit consistent outcomes across these varied architectures. This consistency indicates that our proposed conversion framework is versatile and applicable across numerous neural network models, not limited to traditional structures. It demonstrates the robustness and adaptability of our CASC framework in accommodating different neural network architectures, thereby validating its effectiveness in a broader context.

\begin{table*}[t]
	\centering
		\resizebox{0.9\textwidth}{!}{    %\textwidth, columnwidth
		\begin{tabular}{c|ccccc|ccccc}
			\toprule
			\multirow{2}{*}{Method } & \multicolumn{5}{c}{RegNetX\_8GF} & \multicolumn{5}{c}{ResNeXt50\_32x4d}\\
			\cmidrule{2-11}
			& ANN  & T=8 & T=16  & T=32 & T=64 
			& ANN & T=8 & T=16 & T=32 & T=64\\
			\midrule
			Baseline ($\mathcal{Q}$=8) 
			&70.72  &  1.86 & 18.23 & 36.74 & 42.28
			&  70.40 & 6.99 & 27.39 & 53.20 & 55.80 \\
			Ours ($\mathcal{Q}$=8) 
			& 70.72 & {16.33} & {71.35} & {70.72} & 70.72
			& 70.40  & 21.70 & 71.11 & 70.64 & 70.64\\
			\midrule
			Baseline ($\mathcal{Q}$=16) 
			& 72.63 & 1.52 & 10.71 & 45.28 & 55.82
			& 71.56  & 2.28 & 18.64 & 56.74 & 65.86\\
			Ours ($\mathcal{Q}$=16) 
			& 72.63  & 5.13 & 56.12 & 72.63 & 72.63
			& 71.56  & 7.29 & 56.62 & 71.44 & 71.56\\
			\bottomrule
		\end{tabular}}
		\caption{Conversion accuracy with RegNetX\_8GF and ResNeXt50\_32x4d on the CIFAR100.}
		\label{tab33}
	\end{table*}

	\begin{table}[!t]
		\centering
		\resizebox{0.7\columnwidth}{!}{    %\textwidth, columnwidth
			\begin{tabular}{c|c|c|cccccccc}
				\toprule
				qlevel & RNN & Method & T=2 & T=4 & T=8 & T=16\\
				\midrule
				\multirow{4}{*}{2} & \multirow{4}{*}{\textbf{87.70}} 
				& baseline &  80.51 & 82.96 & 75.36 & 61.72\\
				&    &   baseline+WSC &  80.51 & 87.12 &  87.12 & 87.12\\
				&  &   baseline+CIF &    80.36 & 83.01 & 75.13 & 61.50\\
				&  &  baseline+CIF+WSC & 80.35 & \textbf{87.67}  & \textbf{87.70} & \textbf{87.70}\\
				%			\midrule
				%			\multirow{4}{*}{4} & \multirow{4}{*}{88.17} 
				%			& baseline & 75.25 & 83.38& 86.83 & 87.36 & 87.14\\
				%			&    &   baseline+WSC & 75.25 & 83.38 & 86.83 & 87.88 & 87.88\\
				%			&  &   baseline+CIF &   75.25 & 83.16 & 86.97 & 87.20 & 87.13\\
				%			&  &  baseline+CIF+WSC & 75.25 & 83.16 & \textbf{86.97} & \textbf{88.17} & \textbf{88.17}\\
				\bottomrule
		\end{tabular}}
		\caption{Conversion accuracy comparison on different methods on RNN on the AG NEWS.}
		\label{tab11}
	\end{table}

Furthermore, we systematically discuss the conversion from RNN to SNN. In this study, we have constructed a network composed of an embedding layer, RNN layer, pooling layer, and fully connected layer, which is used for the AG News text classification dataset. We changed the activation function (tanh) of RNN to CQReLU, with a maximum input length of 200. We set all bias parameters to zero during the sleep phase to prevent excessive input. The specific results are shown in  Tab. \ref{tab11}. We compared the baseline with our method and found that for RNN, continuous input would decrease the performance. In contrast, our method achieved the same performance as the RNN, owing to the consistency of the mapping relationship, which demonstrates the potential efficiency of the conversion from RNN to SNN. In future work, how to better combine the advantages of RNN and SNN is worth exploring.

\section{Conclusion}

In this paper, we have conducted a theoretical analysis of the error inherent in the conversion process, focusing on the consistency of the activation function in artificial neural networks. We identified the primary cause of conversion error as the inconsistency in the mapping relationship between the ANN activation function and the total input-output behavior of spiking neurons. To address this, we proposed the CIF neuron and WSC methods, which mitigate the effects of stable points' upper and lower bounds, enabling a nearly loss-free conversion with minimal time steps. Our experimental results demonstrated on complex tasks such as CIFAR100, ImageNet, VOC, and COCO highlight the effectiveness of our method. Notably, our approach equips SNNs with the capability to match the performance of ANNs in intricate tasks, thereby advancing the practical implementation of SNNs in neuromorphic hardware.

However, we also acknowledge the limitations of our current methodology, including the performance declines of ANN due to quantization techniques. In the future, we plan to enhance our approach by incorporating adaptive threshold mechanisms to reduce experimental latency and applying surrogate gradient methods to boost performance. Furthermore, we aim to extend our methodology's applicability to practical scenarios, particularly on FPGAs or neuromorphic hardware. We hope that these future research directions will enhance our understanding of the current work and drive further advancements in spiking neural networks.

\section*{Acknowledgements}

This study is supported by National Key Research and Development Program (2020AAA0107800).

\section*{Author contributions statement}

Li Yang: Conceptualization, Methodology, Software, Resources, Data Curation, Writing - Original Draft.
He Xiang: Investigation, Software, Validation, Writing - Review \& Editing, Visualization.
Kong Qingqun: Investigation, Writing - Review \& Editing, Visualization.
Zeng Yi: Writing - Review \& Editing, Funding acquisition.

\section*{Competing interests} 

The authors declare no competing interests.

 \bibliographystyle{elsarticle-num} 
 \bibliography{sample}

% \end{thebibliography}
\end{document}